\documentclass[journal]{IEEEtran}

\usepackage[graphicx]{realboxes}
\usepackage{xcolor,soul,framed} 
\usepackage{cite}
\usepackage{multirow}
\usepackage{fontawesome}
\colorlet{shadecolor}{yellow}
\DeclareGraphicsExtensions{.pdf,.jpeg,.png}
\usepackage{caption}
\usepackage{subcaption}
\usepackage[cmex10]{amsmath}
\usepackage[colorinlistoftodos]{todonotes}

\usepackage{comment}
\usepackage{enumerate}
\newcommand\textblue[1]{\textcolor{blue}{#1}}
\newcommand\textred[1]{\textcolor{red}{#1}}

\newcommand\quickthings[1]{\textblue{\\\faQuestion #1}}
\newcommand\maybelater[1]{\textred{\\\faClockO#1}}
\newcommand\peter[1]{\textcolor{red}{\faComment #1}}
\newcommand\abs[1]{\\\hl{#1}}

\renewcommand\quickthings[1]{}
\renewcommand\peter[1]{}
\renewcommand\maybelater[1]{}
\renewcommand\abs[1]{}
\renewcommand\hl[1]{#1}

\usepackage[linesnumbered,ruled,vlined]{algorithm2e}

\begin{document}
\bstctlcite{IEEEexample:BSTcontrol}
    \title{ A Comparative Study of Federated Learning Models for COVID-19 Detection  }
  \author{Erfan~Darzidehkalani$^{1}$,
  Nanna M. Sijtsema$^{1}$,
      P.M.A van Ooijen$^{1}$
\\\textit{$^{1}$Machine learning lab, Data Science Center in Health (DASH) \\University of Groningen, Hanzeplein 1, Groningen, The Netherlands}
\\ e.darzidehkalani@rug.nl
  \thanks{ This paper is being prepared with IEEE standards This work was funded in part by NWO under project AMICUS}
  \thanks{Erfan Darzidehkalani is with UMCG (e-mail: e.darzidehkalani@umcg.nl).}
  }
\maketitle
\begin{abstract}

Deep learning is effective in diagnosing COVID-19 and requires a large amount of data to be effectively trained. Due to data and privacy regulations,  hospitals generally have no access to data from other hospitals. Federated learning (FL) has been used to solve this problem, where it utilizes a distributed setting to train models in hospitals in a privacy-preserving manner. Deploying FL is not always feasible as it requires high computation and
network communication resources. This paper evaluates five FL algorithms' performance and resource efficiency for Covid-19 detection. 
A decentralized setting with CNN networks is set up, and the performance of FL algorithms is compared with a centralized environment. We examined the algorithms with varying numbers of participants, federated rounds, and selection algorithms.

Our results show that cyclic weight transfer can have better overall performance, and results are better with fewer participating hospitals. Our results demonstrate good performance for detecting COVID-19 patients and might be useful in deploying FL algorithms for covid-19
detection and medical image analysis in general.

\end{abstract}

\begin{IEEEkeywords}
federated learning, medical image analysis, COVID-19, privacy preserving machine learning
\end{IEEEkeywords}

%
\IEEEpeerreviewmaketitle


\section{Introduction}

\IEEEPARstart{ C}oronaviruses are a family of viruses that cause respiratory and intestinal illnesses in humans and animals. The best-known variants are those responsible for the COVID-19, SARS, and MERS epidemics. Some people with COVID-19 will develop serious complications including COVID pneumonia, which can be recognized in lung CT scans. Research has shown the effectiveness of chest imaging in diagnosing COVID-19-infected people. Deep learning methods, such as Convolutional neural networks (CNN), can help radiologists diagnose COVID-19 with severe symptoms in various image analysis tasks\cite{kogilavani2022covid}.
For Covid-19 deep learning, models have shown great promise in spotting infected areas in CT scans and X-ray images. 

The training of deep learning models requires sufficient and diverse medical datasets gathered from multiple data holders. And most of the existing solutions rely on a central entity in charge of collecting data from different hospitals. However, medical images may contain confidential and sensitive information about patients that often cannot be shared outside the institutions of their origin. One potential solution to this problem is federated deep learning. FL aims to decentralize the whole process of training by keeping the data locally. In FL, the algorithm training is performed in a decentralized manner by different nodes, or clients, that use local data. In this scenario, each decentralized node
trains an individual model using its data and shares the model parameters (instead of the data) with the rest.

FL can differ from centralized data sharing in a number of ways. While both approaches aim to optimize their learning objective, FL algorithms have to account for the fact that communication with clients takes place over unreliable networks with very limited upload speeds. So unlike 
the centralized setting in which computation is generally a bottleneck, in FL communication might be the bottleneck. 

In this paper, we developed a framework that enables collaboration between hospitals and uses multiple data sources to detect COVID-19 infection using FL
The decentralized way of distributing data among different centers guarantees privacy and data is kept locally\cite{darzidehkalanifederatedII}.

\section{Background and Related works}

 
Federated learning has been used for various imaging modalities such as MRI\cite{sheller2020federated} \cite{silva2019federated}, X-ray \cite{balachandar2020accounting} retinal imaging, \cite{balachandar2020accounting}  and for tasks such as brain tumor segmentation \cite{bakas2017advancing} \cite{lee2018privacy} diagnosis \cite{pan2019improving} and treatment 
 selection \cite{lee2018privacy}. FL has shown great promise in developing models to support doctors in making treatment decisions for COVID-19 patients; it was investigated and reported that FL had a clear impact on patient care in a large-scale study on COVID-19 patients across 20 centers on five continents\cite{flores2021federated}. They used chest X-Ray imaging data in addition to clinical data to determine hospital triage for level of care and oxygen requirement in COVID-19 patients. They demonstrated that FL improved model performance for clients with limited datasets, compared to when they were trained on their local data. Another finding was that medical centers with smaller datasets had some classes with only a few patients resulting in underrepresented categories. These clients saw a significant improvement in prediction for those patient categories, which is especially important because, in some clients, less than 5\% of the COVID-19 patients were categorized as having severe symptoms, while more than 95\% had moderate symptoms. However, their care is more critical and requires more attention.

\begin{figure*}[t!]
 \centering
 \includegraphics[width=1\textwidth]{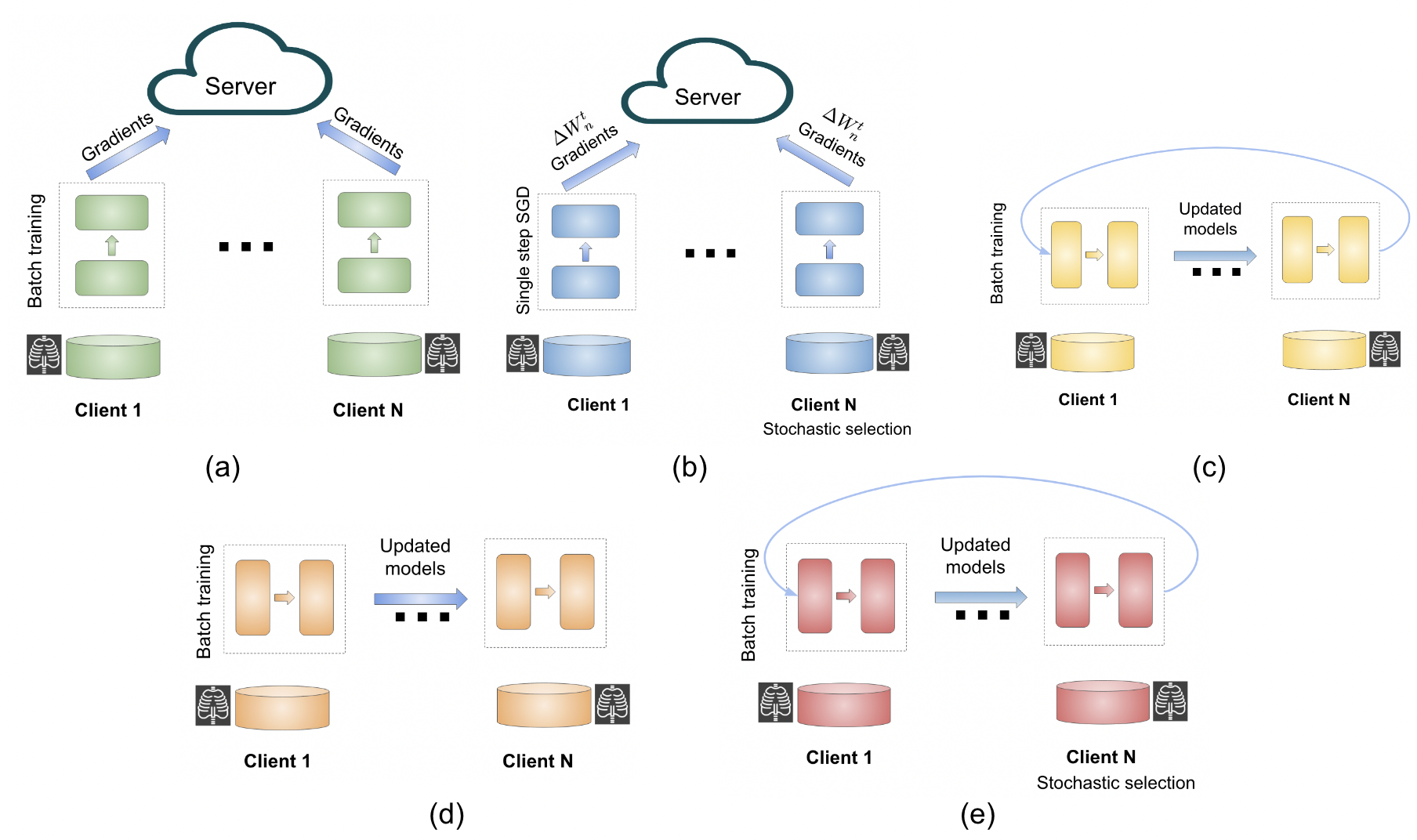}
 \caption{Schematic view of FL models and algorithms, (a) Federated averaging, clients train on local batch of data (b) FedSGD, a subsample of clients are selected, and each performs single step SGD and sends the model updates to the server (c) Cyclic Weight transfer (CWT), clients train locally, and pass the model to the next clients, and the cycle repeats (d) Single weight transfer (SWT) model passes each client only once. (e) Stochastic weight transfer (STWT), the model is passed sequentially through clients, and participating clients in each round are sampled randomly.}
 \label{fig:samplesCTs}
\end{figure*}

Several recent studies have been done to classify scan images of COVID-19 infeced patients, and healthy subjects, and to locate the lesion areas. The primary focus of AI tooling in the management of COVID-19 patients is interpreting radiology images, mainly chest CT, which has been widely applied for detecting lung changes to optimize patient management, and guide treatment decisions\cite{yan2020interpretable}\cite{hu2020challenges}\cite{burian2020intensive}. Other studies have investigated 3D classification networks,  \cite{wang2020weakly} or Covid-19 detection with limited training samples. 
Most of the above studies achieved good accuracy and assumed a centralized environment where one data center has access to all the data.Three publications exist that successfully applied distributed learning for COVID-19 detection\cite{zhang2021dynamic}\cite{kumar2021blockchain}\cite{ho2022fedsgdcovid}.

The global aggregation models used in the above studies were limited to model averaging in federated\cite{ho2022fedsgdcovid}, \cite{zhang2021dynamic}, or blockchain setting \cite{kumar2021blockchain}. Other studies showed limitations of the existing algorithms like a large communication overhead \cite{remedios2020federated}, and problems with convergence or catastrophic forgetting after increasing the number of participating hospitals
\cite{sheller2020federated} \cite{chang2018distributed}.
To our knowledge, no study has been performed that compared multiple FL algorithms under standard conditions to evaluate their applicability. Therefore, comparing multiple FL algorithms under standard conditions could be informative in evaluating their applicability in practice.
To evaluate the existing methods from multiple perspectives, we have implemented the most popular models and compared them in terms of performance, communication overhead, and computation burden.



\section{Algorithms}

\textbf{Centralized data sharing}
In Centralized data sharing (CDS), data is stored in a central location and can be accessed by all clients. This is in contrast to federated and decentralized data sharing methods, where data is stored in multiple locations and accessed by a single user or limited numbers of users. We use CDS as a  baseline  for comparison with other algorithms. 

\textbf{Federated averaging}: The learning procedure for federated averaging is an iterative process containing local and global steps. Each data owner trains a model received from a global server on its local dataset in local iterations \cite{darzidehkalanifederatedI}. The global server updates the global model by aggregating the updated local models. Then it sends it back to clients for the next round.  The optimization problem for federated averaging can be formulated as
\begin{equation}
w^{t+1} = \sum\limits_{i=1}^{N}{p_{i} w_{i}^{t}} , w_{i}^{t}=\arg\min\limits_{w_{i}}{\left(\mathcal{L}(\mathcal{D}_{i};w^{t})\right)}
\end{equation}
where $N$ is the number of data owners, $\mathcal{L}(\mathcal{D}_{i};w^{t})$ is a loss function indicating global model parameters  $w^{t}$ of local datasets, and $p_{i}$ is the probability of selecting client $i$. 
Local optimization can be formulated as $w_{i}^{t+1} \leftarrow w^{t}-\eta\cdot \nabla \mathcal{L}(w^{t};\mathcal{D}_{i})$, where 
 $\eta$ is the learning rate. The global model can be updated based on the local models $w_{i}$ and is shared for aggregation: 
\begin{equation}
w^{t+1} = \sum\limits_{i=1}^{N}{p_{i} w_{i}^{t+1}}
\end{equation}

\textbf{Federated stochastic gradient descent}:Federated Stochastic Gradient Descent (FedSGD) is a variation of Federated Averaging (FedAvg) that uses a large-batch synchronous approach to multi-client learning. FedSGD utilizes a subset of clients from the total number of clients, where $C$ defines the subset of selected clients. This subset of clients is selected at each global round, and the global server sends the most recent global model to them. Each client then performs local training over its dataset for a select number of epochs. The global model is updated based on the local models received from each client and is shared for aggregation, similar to FedAvg. However, in FedSGD, the gradient is computed over the selected batch of clients and therefore, $C<1$, for $C=1$ the training would be non-stochastic (full batch) since all the clients are involved. This allows for training with large batches, as the gradient is computed over the selected subset of clients.   The optimization problem for FedSGD can be formulated as
\begin{equation}
w^{t+1} = w^{t} - \eta \cdot \sum\limits_{i=1}^{C}p_i \cdot \nabla \mathcal{L} (w^t;\mathcal{D}_i)
\end{equation}
where $\eta$ is the learning rate, $p_i$ is the probability of selecting client $i$ and $\mathcal{L}$ is the loss function.
The key difference between FedAvg and FedSGD lies in the use of large-batch synchronous approach in FedSGD. This approach has been shown to outperform the naive asynchronous SGD training due to the increased accuracy and efficiency, as compared to the local training approach used in FedAvg \cite{chai2020fedeval}\cite{charles2021large}. Additionally, FedSGD has been shown to be more robust to non-IID data distributions, compared to FedAvg \cite{chai2020fedeval}. 
\quickthings{This part is unclear to me.}
\maybelater{paraphrised}

\textbf{Cyclic weight transfer}: Federated learning techniques have been widely used in medical image processing tasks using a method known as cyclic weight transfer (CWT)\cite{balachandar2020accounting}. This method involves training models on individual clients for a number of iterations and then cyclically sharing the updated weights with the following client. However, the existing CWT algorithm faces a notable challenge, as it lacks the ability to effectively manage inter-client variability in training data or labels.  To ensure the practical application of CWT, it is crucial to develop a version that can handle the common variations observed in a majority of real-world medical imaging datasets.\cite{darzidehkalanifederatedII}

\textbf{Single weight transfer}: Single weight transfer (SWT) is another FL model widely used in the medical imaging domain. In Single weight transfer, models are trained in each client with its local data, and then the updated model is transferred to the next client. The difference between this method and CWT is that here the model passes each client only once. 

\textbf{Stochastic weight transfer}:
In stochastic weight transfer (STWT), we select a subsample of clients and train them in a cyclic manner. Similar to FedSGD, a ratio defines the number of selected clients to the total number of clients in each federated round.





\section{Experiment}
\textbf{Dataset} Our experiments used two publicly available data sources, the Tongji hospital dataset and Brazil's SARS-CoV-2 dataset. Tongji dataset consists of 349 chest CT-scans of COVID-19 positive and 397 scans of healthy subjects, all low-resolution CT modalities. Brazil's SARS-CoV-2 dataset consists of 2482 samples, 1252 scans of COVID-19-infected patients, and 1230 healthy subjects collected from multiple hospitals in Sao Paulo, Brazil. Train and test sets were obtained randomly from the aggregated datasets. Table \ref{table_datadist}, shows data distribution.

\begin{table}[h!]
\centering
\setlength{\tabcolsep}{6pt}
\renewcommand\arraystretch{1.22}
\caption{ \small Data distribution}
\begin{tabular}{| *{5}{c|} }
\hline
Data  & Class & Dataset & Train & Test
\\   \hline  
\multirow{2}{3em}{Covid}     &Brazil&1252&\multirow{2}{2em}{1451}  &\multirow{2}{2em}{150}  \\
&Tongji&349&&  \\
\hline
\multirow{2}{5em}{Non-Covid}     &Brazil&1230&\multirow{2}{2em}{1477}  &\multirow{2}{2em}{150}  \\
&Tongji&397&&  \\
\hline
\end{tabular}
\label{table_datadist} 
\end{table}

\textbf{Preprocessing}
Images were selected as 2D slices in greyscale. Preprocessing included randomly cropping between 0.5 to full size, random horizontal flipping, and intensity normalization. CT-slices were all resized to 224$\times$224 pixels with interpolation. Figure \ref{fig:samplesCTs} shows samples of processed images.
\begin{figure}[h!]
 \centering
 \includegraphics[width=0.45\textwidth]{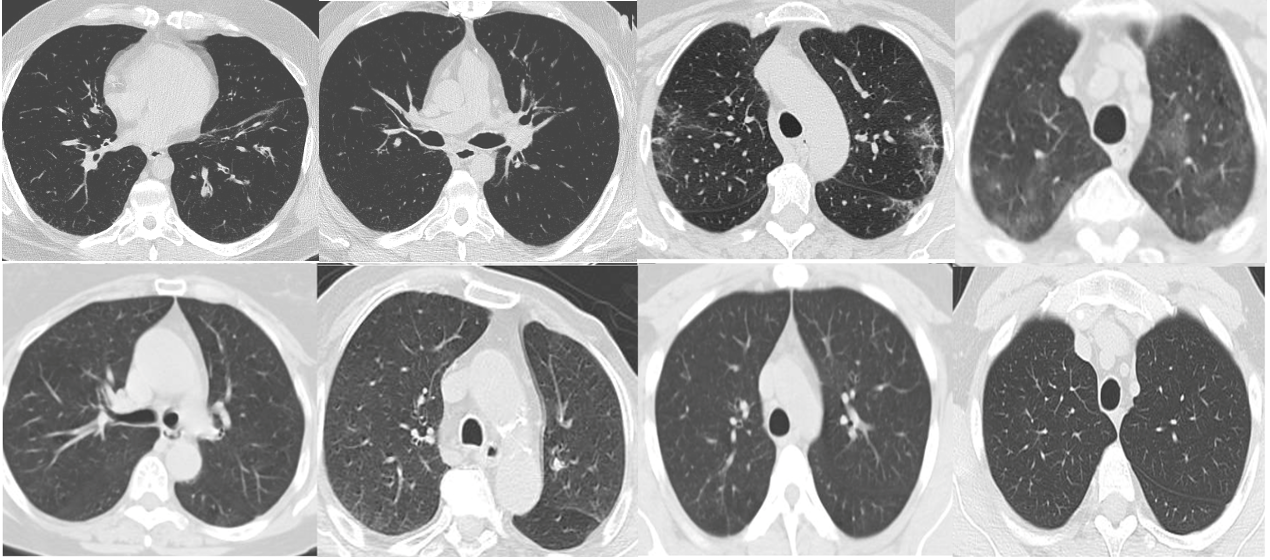}
 \caption{Sample CT slices of Covid-19 images (top row) and Non-covid images (bottom row)}
 \label{fig:samplesCTs}
\end{figure}


\textbf{Training}
ResNet-18 is used as the backbone deep learning model. ResNet-18 comprises one initial block cascaded to four middle blocks. The initial block is made of convolutional,  batch normalization, ReLU, and pooling layers. Middle blocks have the same layers, connected with straight and skip connections. The model is pre-trained on ImageNet dataset \cite{he2016deep} with a CrossEntropy loss function and learning rate of $0.05$. 
Each federated round consisted of 20 internal epochs for each client and batches of 16 samples in each iteration. For models which use minibatch training, like STWT and FedSGD, a subset of clients is randomly selected. Similar to training, test data was split into mini-batches, and the results were averaged across batches. We performed training with various participating clients and federated rounds to evaluate their effect on final performance. Models were also trained in a centralized, non-federated setting to build a comparison baseline.

\begin{figure}[h!]
 \centering
 \includegraphics[width=0.50\textwidth]{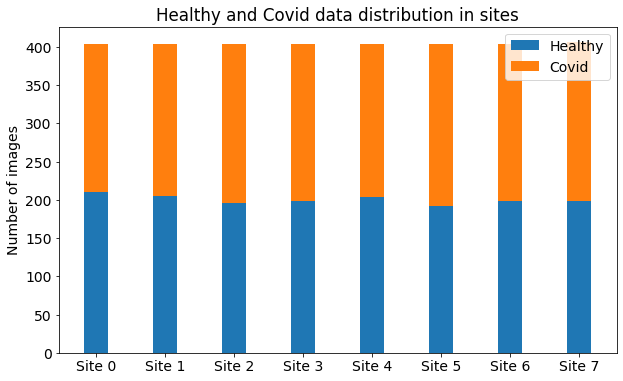}
 \caption{Data distribution of each client in the simulated federated setting}
 \label{fig:pgd-atta-comparison}
\end{figure}

\textbf{Evaluation} 
Standard classification metrics, accuracy, recall, precision, and F1 score, were used as our evaluation criteria. We also evaluated the level of communication, the amount of transferred data in each algorithm, and the computational complexity of the models.

\label{sec:experiment}

\section{Results}


Here, the result for the setting with 10 participating clients and a maximum of 10 rounds is presented. The results are average performance among clients for all the federated rounds. Table \ref{best_performance} shows the results.


\begin{table}[h!]
\centering
\setlength{\tabcolsep}{5.5pt}
\renewcommand\arraystretch{1.12}
\caption{ \small Comparison of FL algorithms on classification of COVID-19 data for 10 clients, averaged performance in all the 10 rounds.}
\begin{tabular}{| *{5}{c|} }
\hline
Method & Accuracy & Recall & Precision & F1 score
\\
  \hline

CDS
 &87.75\%&89.57\% &87.93\% & 87.19\%  
\\   \hline  
FedAVG
 & 66.72\% &70.02\% &43.80\% & 51.7\%\\ 
 \hline  
FedSGD
 &65.17\% &68.24\%&43.86\%& 47.75\%  
 
 \\
 \hline  
CWT
 &87.75\%&89.00\%& 88.67\%&87.52\%\\
 
   \hline

SWT
 & 64.60\% &74.33\% & 65.55\% &59.66\% \\
    
\hline 
STWT
 & 84.21\% &84.09\%&83.33\% & 81.71\%  \\

\hline
\end{tabular}
\label{best_performance} 
\end{table}

\textbf{Effect number of federated rounds}

To evaluate the effect of number of rounds, models with 3, 5, 10 and 15 rounds were tested. The test results are shown for both centralized and FL  algorithms. Table \ref{number_of_rounds} shows the results of our experiment. The increasing number of rounds correlates with higher accuracy of the global model.







\begin{table}[h!]
\centering
\setlength{\tabcolsep}{3.5pt}
\renewcommand\arraystretch{1.12}
\caption{ \small Effect number of rounds on accuracy of FL algorithms for 10 clients, 20 internal epochs.}
\begin{tabular}{| *{5}{c|} }

\hline

Method & 3 rounds & 5 rounds & 10 rounds & 15 rounds
\\
 \hline
CDS
 &85.06\% &81.56\% & 91.06\% & 91.04\% 
\\
\hline
FedAVG
 & 56.05\%  &63.78\%&69.64\%&70.73\% \\

 \hline
 FedSGD
 &  50.88\% & 55.9\% &75.59\% & 76.94\%\\
\hline  
CWT
 &  80.77\% & \textbf{89.78\%} &\textbf{91.27\%}& \textbf{93.56\%}\\


\hline  
STWT
 &  \textbf{90.73\%} & 83.97\% &89.44\% & 93.01\%\\

  \hline

\end{tabular}
\label{number_of_rounds} 
\end{table}
\quickthings{Shouldn't this column have the same results as table II for accuracy?}

\peter{Table III and Table II do not represent the same thing. Table II is average performance in all 10 rounds, is to give an overall impression of accuracy, and table III is the final result in the end of the 5th, 10th , 15th rounds.  Added  to the table caption.
}

\begin{figure}[h!]
 \centering
 \includegraphics[width=0.50\textwidth]{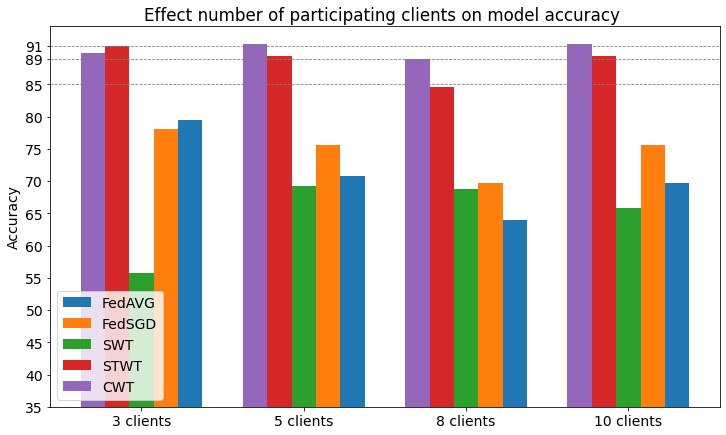}
 \caption{Accuracy of FL algorithms with differrent number of clients. }
 \label{fig:data distribution}
\end{figure}

\textbf{Effect number of participating clients}
To evaluate the number of clients on the FL network, we examined scenarios with 3,5, and 8 participating clients. We trained each of the clients in 20 internal epochs. The number of Federated rounds for all the algorithms (except SWT) was 10.
The average test results are shown in the Figure \ref{fig:data distribution}. \quickthings{Why not a column for 10 clients here?}
\maybelater{ 10 clients are shown in table three , third column. Added 10 clients here as well}






\begin{figure}[h!]
 \centering
 \includegraphics[width=0.50\textwidth]{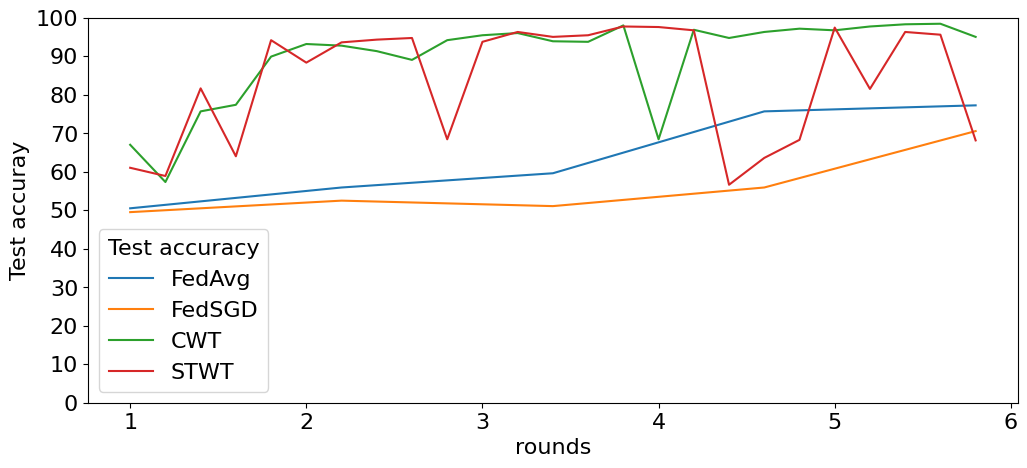}
 \caption{Test accuracy as a function of passing rounds}
 \label{seq-vs-noseq}
\end{figure}

\begin{table}[h!]
\label{computation time}
\centering
\setlength{\tabcolsep}{7pt}
\renewcommand\arraystretch{1.22}
\caption{ \small Computation time (seconds) for FL algorithms for standardized setting}
\begin{tabular}{| *{9}{c|} }
\hline
Method & 3 clients & 5 clients & 8 clients  & 10 clients
\\   \hline  
FedAVG
 & 8934 sec&8975 sec  & 9002 sec & 9030 sec\\

\hline  
FedSGD
 & 8810 sec &8853 sec& 9013 sec & 9052 sec \\
\hline  
CWT
 &  5119 sec& 5450 sec&5383 sec & 5556 sec\\
 \hline
STWT &
2805 sec&  5243 sec& 6101 sec & 6129 sec\\
 \hline  
SWT
 & \textbf{543 sec} &\textbf{547 sec} & \textbf{589 sec } & \textbf{618 sec}\\


\hline  

\end{tabular}
\label{table_time} 
\end{table}

Communication can also be a bottleneck in this setting. In methods like federated averaging, the lower bounds for total communicated data are proportional to $\sim$
${2NT}$
 where T is the number of rounds and N is the number of participating clients. In CWT, this lower bound is  $\sim$ ${NT}$. In our setting, we use a ResNet 101 model. We calculated the overall transferred data for the different number of rounds. As expected, in settings where clients are chosen stochastically, the communication time is lower than full-client participation. \hl{Also, the experiments for computational costs show that non-sequential models generally have higher computational requirements than sequential models.}
\quickthings{Again, why only this comparison is relevant enough to put into the text?}
\maybelater{Here fedavg, and fedsgd are the same family, (non sequential), and CWT are the other family (sequential)., Changed the phrasing from CWT/FedAVG to seq/non-seq, hope it's okay this time.}

\begin{table}[h!]
\centering
\label{Transferred data}
\setlength{\tabcolsep}{5.5pt}
\renewcommand\arraystretch{1.12}
\caption{ \small Comparison of total transferred data in a normalized setting in (GB)}
\begin{tabular}{| *{5}{c|} }
\hline
Method & 3 rounds & 5 rounds & 10 rounds & 15 rounds 
\\   \hline  
FedAVG
 & 1.371 &2.286 & 4.571 & 6.857\\

 \hline
 FedSGD
 &  0.823 & 1.371 &2.743 & 4.114\\
\hline  
CWT
 &  0.686 & 1.143 &2.286 & 3.428\\


\hline  
STWT &
  0.411 & 0.686 &1.371 & 2.057\\

 \hline
\end{tabular}
\label{table_time} 
\end{table}


\label{sec:results}


\section{Discussion}
\label{sec:discussion}





%
Our results show that FL has comparable performance to centralized data sharing, with the advantage of keeping data private. With large volumes of data and after high number of rounds, centralized data sharing and cyclic weight transfer have the highest accuracy.

Sequential models are susceptible to catastrophic forgetting, where a global model performs well on the latest client it has seen while having poor performance in other clients.\quickthings{cryptic sentence}\maybelater{I have paraphrised the sentence}
Conversely, in algorithms like FedAvg and FedSGD, the models are averaged asynchronously after all the clients have finished their training. So the trajectory is smoother and overall improving with more communication rounds. As shown in Figure \ref{seq-vs-noseq}, local test results can have a high variance when passing through clients sequentially, indicating the catastrophic forgetting effect.

Models like FedAvg, and FedSGD, in which all the clients have identical copies of one global model, are slower and more challenging to converge compared to sequential models like CWT and STWT. Also, FedAvG and FedSGD require more training resources due to active server participation, resulting in more computation and network consumption. Stochastic client selection is an efficient way of training. Stochastic models save significant time and resources while having similar performance to full client participation. Overall, CWT and STWT have best results in terms of model accuracy and computation times. These findings could be practical in further federated deployments in medical institutions.



 

Sequential models like CWT and STWT perform better than non-sequential models on fewer training rounds. For example,  after three rounds of training, STWT and CWT both reach 96\% accuracy, while FedAvG reaches 66\%, and FedSGD performs equally to a random classifier. As the training proceeds, FedAVG and FedSGD gradually improve with more global rounds.The concept of sequential models is similar to fine-tuning \cite{chen2020online} in centralized deep learning, so in cases where a hospital temporarily joins an FL network, or there is an urgency in training, sequential models are a better option.

More training rounds do not always lead to a better global model. Although average performance on all clients improves, more global rounds lead to worse performance for some clients. The global model can overfit some clients, leading to lower performance on others\cite{mohri2019agnostic}. Some studies suggested early stopping and fine-tuning to local dataset after global training is finished \cite{yu2020salvaging}. 
In all the algorithms, more clients resulted in slower convergence. This effect is stronger in the FedAvg algorithm. In FedAVG, the Global model must compromise between potentially disparate local minima.\cite{li2019convergence} Methods such as adaptive or stochastic selection of clients and momentum-based models help faster convergence. \cite{liu2020accelerating}
Our results suggest that stochastic client participation is close to full client participation. The average results of four trials with varying rounds, shown in Table \ref{number_of_rounds} indicate that stochastic client participation in FedSGD results in 5.23\% performance loss and 40\% less bandwidth consumption compared to FedAvg. In STWT, it results in only 1.25\% less accuracy but saves 40\% of communication and 11.3\% of computation.
These results are in accordance with prior studies, showing that, in theory, stochastic and full client participation have similar global minima\cite{cho2020client}. Stochastic client selection can be advantageous when there are limited resources, or in larger networks with occasionally unavailable clients.




We did not assume any shift in clients' data. A more comprehensive analysis should consider the effect of the domain and distribution shifts on the performance of the algorithms. Also, inter-client data variability and the effect of heterogenous clients could be a future line of research.

\section{Conclusion}
\label{sec:conclusion}
 
FL enables extensive collaborations of hospitals to address medical imaging problems while keeping data private. Real-world implementation requires consideration of efficiency and hardware requirements in addition to model performance, especially in the healthcare field, which generally has limited infrastructure. We implemented five FL algorithms for COVID-19 detection and analyzed their efficiency and accuracy.
Our results suggest that FL algorithms have comparable performance to centralized data sharing, with the advantage of keeping data private. They also show that the sequential methods are a better option in most of the scenarios. This study can be helpful in the deployment of FL systems in COVID-19 detection and medical image analysis in general.


\section*{Acknowledgement}
This research is supported by KWF Kankerbestrijding and the Netherlands Organisation for Scientific Research (NWO)  Domain AES, as part of their joint strategic research programme: Technology for Oncology IL. The collaboration project is co-funded by the PPP allowance made available by Health Holland, Top Sector Life Sciences \& Health, to stimulate public-private partnerships. The authors acknowledge Nikos Sourlos for his comments.

\ifCLASSOPTIONcaptionsoff
  \newpage
\fi

\bibliographystyle{IEEEtran}
\bibliography{IEEEabrv}

\vfill
\end{document}